\documentclass[10pt,twocolumn,letterpaper]{article}

\usepackage[pagenumbers]{cvpr} 

\usepackage{graphicx}
\usepackage{amsmath}
\usepackage{amssymb}

\usepackage{array,booktabs,tabularx,multirow}
\newcolumntype{Y}{>{\centering\arraybackslash}X}
\usepackage{caption}
\usepackage{subcaption}
\usepackage{graphicx}
\usepackage{fontawesome}
\usepackage[table]{xcolor}
\usepackage{float}
\usepackage{ifthen}
\usepackage{etoolbox}

\usepackage{algorithm}
\usepackage[noend]{algpseudocode}
\usepackage{eqparbox}

\algrenewcommand\alglinenumber[1]{\footnotesize #1}

\definecolor{highlightrow}{rgb}{0.92, 0.92, 0.98}

\newcommand{\tablestyle}[2]{\setlength{\tabcolsep}{#1}\renewcommand{\arraystretch}{#2}\centering\footnotesize}
\newlength\savewidth
	
\newcommand{\etalcite}[2]{#1 \emph{et al.} \cite{#2}}

\newcommand{\thesiscitelong}{%
	\ifdef{\reviewversion}{%
		\etalcite{Authors}{anon-thesis}%
	}{%
		\etalcite{Pyrr{\"o}}{pasi-thesis}%
	}%
}%

\newcommand{\repolink}{%
	\ifdef{\reviewversion}{%
		\texttt{file:///path/to/air.zip}%
	}{%
		\url{https://github.com/Accenture/AIR}%
	}%
}%

\newcommand{\uniname}{%
	\ifdef{\reviewversion}{%
		Unknown University%
	}{%
		Aalto University%
	}%
}%

\newcommand{\thesiscite}{%
	\ifdef{\reviewversion}{%
		\cite{anon-thesis}%
	}{%
		\cite{pasi-thesis}%
	}%
}%
	
\makeatletter
\newcommand\rowlabel[2]{#1\def\@currentlabel{#1}\label{#2}}
\makeatother

\usepackage[pagebackref,breaklinks,colorlinks]{hyperref}

\hypersetup{
	colorlinks,
	citecolor=black,
	linkcolor=black
}

\usepackage[capitalize]{cleveref}
\crefname{section}{Sec.}{Secs.}
\Crefname{section}{Section}{Sections}
\Crefname{table}{Table}{Tables}
\crefname{table}{Tab.}{Tabs.}

\def\cvprPaperID{9402} 
\def\confName{CVPR}
\def\confYear{2022}

\newcommand{\eqvskip}{\vspace{-2ex}}

\begin{document}

\title{Rethinking Drone-Based Search and Rescue with Aerial Person Detection}

\author{
	Pasi Pyrrö\\
	Accenture, Aalto University\\
	{\tt\small pasi.pyrro@accenture.com}
	\and
	Hassan Naseri\\
	Accenture\\
	{\tt\small hassan.naseri@accenture.com}
	\and
	Alexander Jung\\
	Aalto University\\
	{\tt\small alex.jung@aalto.fi}
}
\maketitle

\begin{abstract}
	The visual inspection of aerial drone footage is an integral part of land search and rescue (SAR) operations today. Since this inspection is a slow, tedious and error-prone job for humans, we propose a novel deep learning algorithm to automate this aerial person detection (APD) task. We experiment with model architecture selection, online data augmentation, transfer learning, image tiling and several other techniques to improve the test performance of our method. We present the novel Aerial Inspection RetinaNet (AIR) algorithm as the combination of these contributions. The AIR detector demonstrates state-of-the-art performance on a commonly used SAR test data set in terms of both precision (${\sim}$21 percentage point increase) and speed. In addition, we provide a new formal definition for the APD problem in SAR missions. That is, we propose a novel evaluation scheme that ranks detectors in terms of real-world SAR localization requirements. Finally, we propose a novel postprocessing method for robust, approximate object localization: the merging of overlapping bounding boxes (MOB) algorithm. This final processing stage used in the AIR detector significantly improves its performance and usability in the face of real-world aerial SAR missions.
	
	
\end{abstract}

\let\thefootnote\relax\footnotetext{\noindent This paper is based on a master's thesis work from \uniname \thesiscite. Code is available at \repolink.}

\section{Introduction}
\label{sec:intro}

Search and rescue (SAR) missions have been carried out for centuries to aid those who are lost or in distress, typically in some remote areas, such as wilderness. With recent advances in technology, small unmanned aerial vehicles (UAVs) or drones have been used during SAR missions for years in many countries \cite{Bozic-Stulic2019, Goodrich2008, Marusic2019, Rudol2008, Gotovac2020}. The reason is that these drones enable rapid aerial photographing of large areas with potentially difficult-to-reach terrain, which improves the safety of SAR personnel during the search operation. Moreover, drone units can outperform even several land parties in search efficiency \cite{Karpowicz2016}. However, there remains the issue of inspecting a vast amount of aerial drone images for tiny clues about the missing person location, which is currently a manual task done by humans in most cases. It turns out this inspection process is very slow, tedious and error-prone for most humans, and can significantly delay the entire drone search operation \cite{Bozic-Stulic2019, Rudol2008, Marusic2019}.

\begin{figure}[t]
	\centering
	\captionsetup[subfigure]{labelfont=bf}
	\begin{subfigure}{0.31\linewidth}
		\centering
		\includegraphics[width=\linewidth]{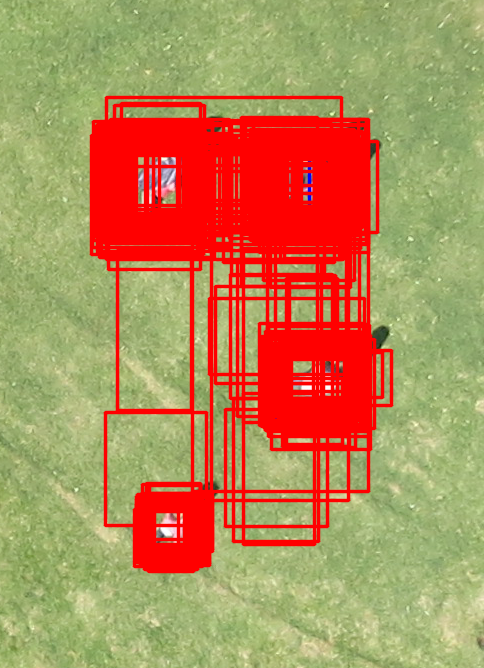}
		\caption{Input boxes}
		\label{fig:no-bbox-aggregation}
	\end{subfigure}%
	\hfill
	\begin{subfigure}{0.31\linewidth}
		\centering
		\includegraphics[width=\linewidth]{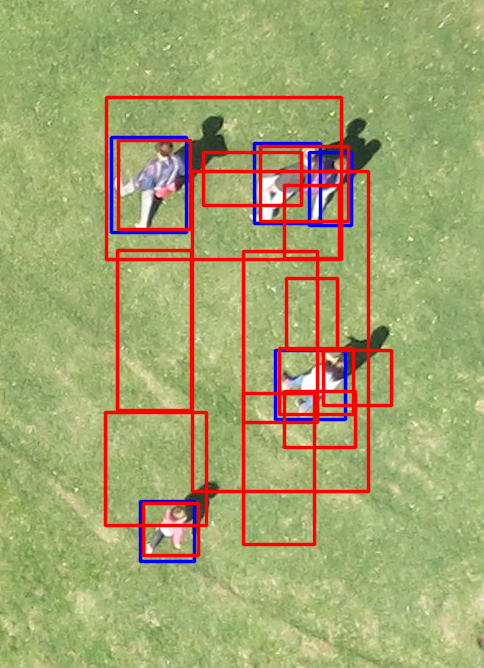}
		\caption{NMS results}
		\label{fig:nms}
	\end{subfigure}%
	\hfill
	\begin{subfigure}{0.31\linewidth}
		\centering
		\includegraphics[width=\linewidth]{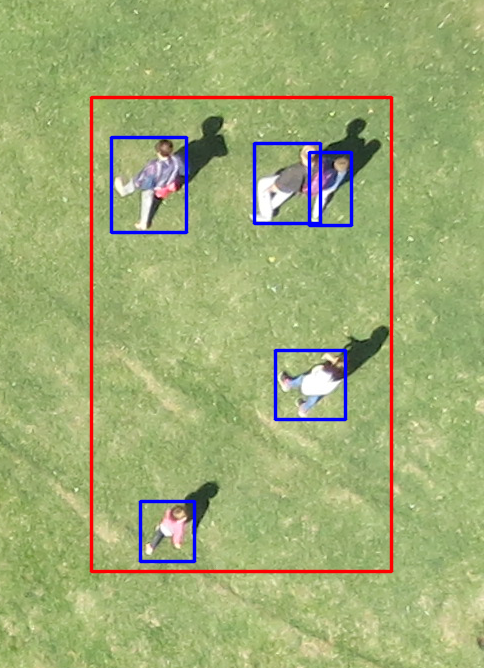}
		\caption{MOB results}
		\label{fig:mob}
	\end{subfigure}
	\vspace{2ex}
	\caption{Ground truth labels (blue) and three different detection results (red) from the AIR object detector: (a) raw, non-aggregated bounding boxes, (b) NMS-postprocessed boxes, and (c) MOB-postprocessed boxes. In typical object detection scenarios (e.g., the PASCAL VOC2012 challenge \cite{Everingham2011}), one tries to achieve very accurate boxes that result in (b), although (c) results frequently suffice in SAR-APD tasks. Moreover, (c) results are visually more pleasing and much easier to produce. Indeed, in this paper we analyze detection and evaluation methods that encourage results similar to (c) that overlook the difficult and irrelevant \textit{dense group separation problem} in SAR-APD. It could be argued that producing and rewarding (c) results better reflects the true objective of aerial SAR searches, which is to provide the ground search teams with an approximate location of clues about the missing person.}
	\label{fig:bbox-aggregation}
\end{figure}

\begin{figure}[t]
	\centering
	\includegraphics[width=\linewidth]{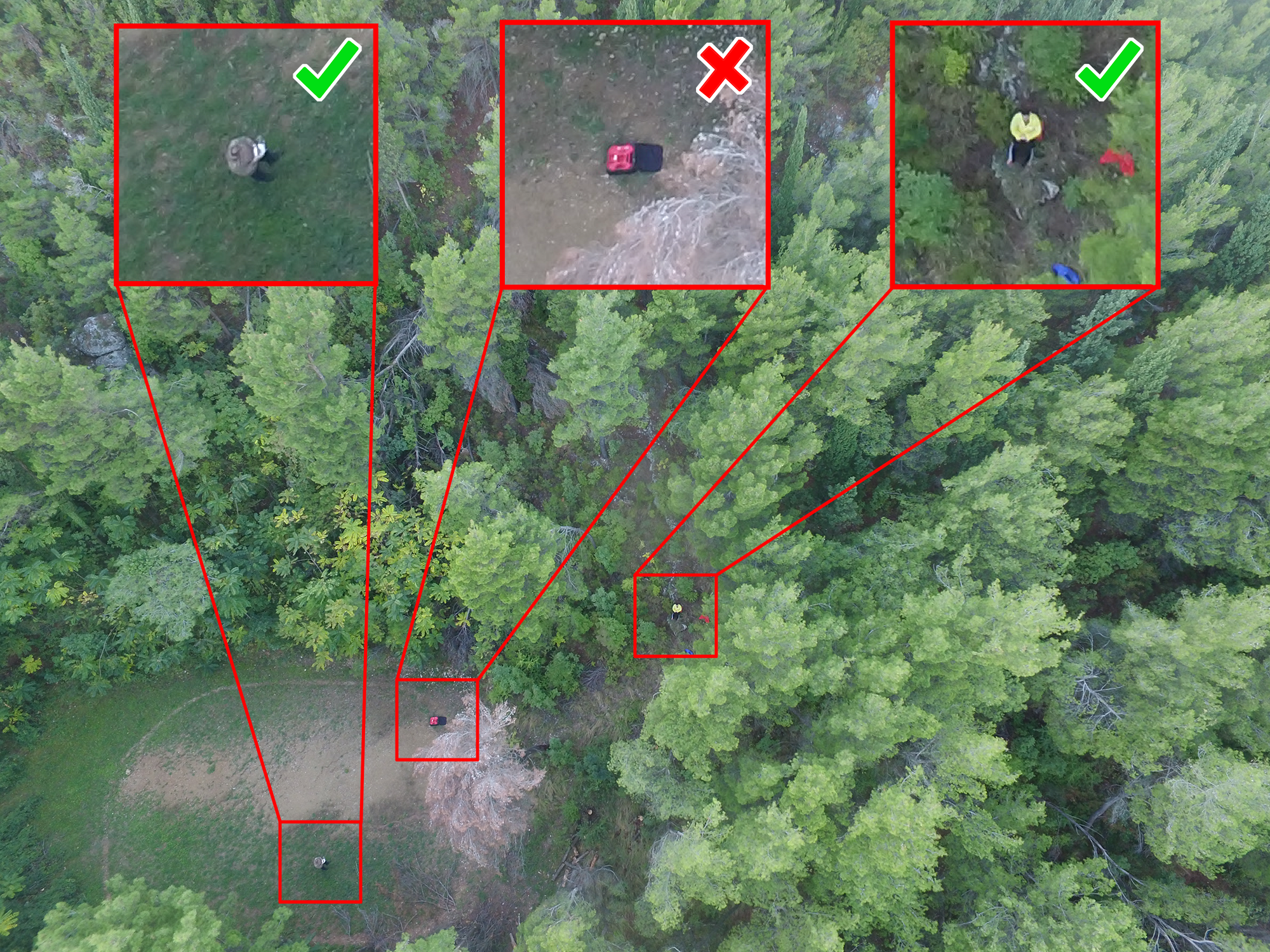}
    \vspace{-2.5ex}
	\caption{An example of visual search problem in SAR context. As evident
		from the image, it is really difficult to spot small humans without distinctive
		colors from still, high-resolution images. The aerial image is taken from the HERIDAL data set \cite{Bozic-Stulic2019}.}
	\label{fig:visual-search}
\end{figure}

In this paper we propose a novel object detection algorithm, called \textit{Aerial Inspection RetinaNet} (AIR), that automates the visual search problem of aerial SAR missions, which is illustrated in \cref{fig:visual-search}. The AIR detector is based on the novel RetinaNet deep learning model by \etalcite{Lin}{Lin2017Retina}. RetinaNet is well-suited for addressing the imbalance of foreground and background training examples \cite{Lin2017Retina}, which can be particularly severe in aerial image data. Furthermore, AIR incorporates various modifications on top of RetinaNet to improve its performance in \textit{small object detection} from aerial high-resolution images. Consequently, AIR achieves state-of-the-art performance on the difficult HERIDAL benchmark \cite{Bozic-Stulic2019} both in terms of precision (${\sim}$21 percentage point increase) and processing time while achieving competitive recall. This commendable performance is achieved through multiple strategies, such as selecting an appropriate \textit{convolutional neural network} (CNN) backbone, conducting an ablation study and calibrating the model confidence score threshold.

Moreover, we reformulate the evaluation criteria for the problem coined \textit{search and rescue with aerial person detection} (SAR-APD). This problem is associated with the used HERIDAL data set consisting of aerial search imagery in SAR setting (see \cref{fig:visual-search}). The reason for this reformulation is that the HERIDAL benchmark itself does not define clear evaluation criteria for SAR-APD methods, such that different solutions could be compared reliably. Hence, we define two algorithms for ranking this benchmark problem: \textit{VOC2012} \cite{Everingham2011} and the proposed \textit{SAR-APD evaluation scheme}. We recommend using the latter for computing the main performance metrics of the HERIDAL benchmark because it focuses on detection rather than localization. This scoring trade-off is designed for SAR-APD problems, since reliable detection is more important in real-world SAR operations. Nevertheless, we suppose the conventional VOC2012 evaluation can be used for computing additional ``challenge'' metrics on HERIDAL. This could be beneficial for non-SAR-related \textit{aerial person detection} (APD) applications using the HERIDAL data, which might require counting the persons for example. Ultimately, our proposed evaluation scheme is better aligned with the real-world SAR operational requirements for localization, which is typically tens of meters in ground resolution \cite{Molina2012}.

Furthermore, we deem the traditional \textit{non-maximum suppression} (NMS) \cite{nms} bounding box aggregation (BBA) algorithm, which is used in many modern object detectors to postprocess detection results \cite{Bodla2017}, insufficient for solving the \textit{dense group separation problem} of HERIDAL test data. NMS is designed for exact delineation of individual objects, however, this is a daunting task for the small person groups of HERIDAL, as illustrated in \cref{fig:nms}. Therefore, we propose a different BBA strategy that merges the overlapping bounding box clusters, as shown in \cref{fig:mob}. This \textit{merging of overlapping bounding boxes} (MOB) algorithm is well-suited for approximate, robust localization and detection of small objects in SAR-APD tasks. MOB is primarily designed to be used with our SAR-APD evaluation scheme for testing object detector performance (i.e., not currently used during training). Moreover, both the proposed algorithms are independent from AIR and straightforward to implement into most object detector frameworks.

The rest of the paper is organized as follows. In \Cref{sec:sar-apd} we explain the SAR-APD problem and highlight some related work on it. Moreover, we formalize the evaluation criteria for ranking object detector performance in SAR-APD, and define our novel approach to that. \Cref{sec:mob} describes the proposed MOB postprocessing algorithm. In \Cref{sec:air} our AIR detector is briefly described, and \Cref{sec:experiments} discusses our experiment results with the state-of-the-art comparison. Finally, \Cref{sec:conclusions} concludes the paper.


\section{Aerial person detection problem in SAR}
\label{sec:sar-apd}

Aerial person detection is a very specific class of object detection problems, which involves detecting people from aerial perspective. It is a difficult problem in general due to various challenges, such as rapid platform motion, image instability, varying environments, high variability in viewing angles and distances, and extremely small size of objects. In SAR context, additional constraints are included to the APD problem: straight-down (nadir) image viewpoint (see \cref{fig:visual-search}), 4K resolution for images and a relatively high drone flying altitude (from 40 to 65 m). These requirements are embedded in the HERIDAL data and are based on real-world SAR experience \cite{Bozic-Stulic2019}. For example, this nadir viewpoint minimizes occlusion by vegetation and simplifies georeferencing of the image observations. Additionally, the SAR-APD problem definition emphasizes successful detection over accurate localization because fine-grained search can be carried out by ground teams dispatched to the clue location \cite{Goodrich2008, Molina2012}.

\subsection{Related Work}

Despite being a relatively new and very specific research field, many works have been proposed in SAR-APD. Since the field is still lacking a comprehensive theoretical foundation and methodology, these approaches vary significantly. The SAR-APD methods range from simple thresholding rules for nearly raw pixel values \cite{Sun2016} to complicated deep learning systems \cite{Marusic2018, Bozic-Stulic2019, Vasic2020}. Due to the many recent successes of deep neural networks in image processing \cite{Krizhevsky2009, Szegedy2015, Long2015, He2016, Ren2016, Liu2016, Hu2018}, our detection algorithm also falls into the latter category.

An intuitive solution is to use infra-red cameras in UAVs to distinguish the human thermal signature from background \cite{Rudol2008, Gaszczak2011}. Nevertheless, this approach has many limitations, especially during summer in rocky areas when the background thermal radiation can cloak that of humans \cite{Bozic-Stulic2019}. Furthermore, high-resolution thermal cameras can be too expensive for typical SAR efforts with limited budget.

The mean shift clustering algorithm by Comaniciu and Meer \cite{Comaniciu2002} has seen extensive use in SAR-APD \cite{Turic2010, Sokalski2010, Gaszczak2011}. This general, nonparametric clustering method enables straightforward segmentation of distinctively colored regions and works adequately well in practice. However, the lack of supervised training with diverse data limits the generalizability of this approach (e.g., if the person is not wearing distinctively colored clothing).

Salient object detection is another SAR-APD research direction \cite{Gotovac2016, Bozic-Stulic2019}, which attempts to extract visually distinct regions that stand out from the input images. This is inspired by the human visual ability \cite{Borji2019}. The wavelet transform based saliency method \cite{Imamoglu2012} has shown promising results for proposing class-agnostic candidate object regions (e.g., 93\% recall on HERIDAL test data \cite{Bozic-Stulic2019}). Nevertheless, the individual region classification to eliminate the vast amount of false positives remains a problem, since it is computationally expensive, and the classifier training with cropped image patches can be cumbersome.

Deep learning solutions to SAR-APD have been studied extensively as well. \etalcite{Maru{\v{s}}i{\'{c}}}{Marusic2018} used the well-known Faster R-CNN detector \cite{Ren2016}, which achieved commendable performance on HERIDAL with little modification. \etalcite{Bo\v{z}i{\'{c}}-\v{S}tuli{\'{c}}}{Bozic-Stulic2019} used a VGG16 CNN classifier on top of their salient region proposals. Lastly, Vasi\'{c} and Papi\'{c} \cite{Vasic2020} proposed a two-stage multimodel CNN detector, which uses an ensemble approach of two deep CNNs for both region proposal and classification stages in the detector pipeline. This complex model achieves a very high recall on HERIDAL test data at the cost of notably reduced inference speed.

\subsection{HERIDAL data set}

The HERIDAL data set \cite{Bozic-Stulic2019} is currently the only widely used, public SAR-APD benchmark available as far as we know. It is constructed using expert knowledge and statistics compiled by Koester \cite{koester2008lost} and the Croatian SAR team to ensure proper coverage of realistic SAR scenarios, according to \etalcite{Bo\v{z}i{\'{c}}-\v{S}tuli{\'{c}}}{Bozic-Stulic2019}. This data set contains 1647 annotated high-resolution UAV aerial images taken from Mediterranean wilderness of Croatia and BiH, that is, 1546 training and 101 testing images. We use 10\% of these training samples for validation. An example image of the data set is shown in \cref{fig:visual-search}. Furthermore, the data set contains only a single object class ``person", instances of which are annotated with bounding boxes using the VOC XML label format \cite{Everingham2010}. Each image in HERIDAL contains at least one person annotation, and some images contain more than ten. In total, HERIDAL contains 3229 person annotations (2892 train and 337 test).

The high-resolution aerial images are taken from several locations with various drone and camera models at altitudes ranging from 40 to 65 m. Most of the images have $4000 \times 3000$ pixel resolution, however, a few pictures have a wide-format-resolution of $4000 \times 2250$ pixels. 
In other words, the aerial images have an approximate 2 cm ground resolution. That is, one pixel width corresponds to roughly 2 cm on the ground \cite{Bozic-Stulic2019}, which is rather accurate.

However, all objects in the images are very small, as shown in \cref{fig:visual-search}. Indeed, the objects have approximate dimensions of $60 \times 60$ pixels on average, which is only 0.03\% of the whole image area. This small object size poses a considerable challenge for object detection methods \cite{Liu2020}.

\subsection{Evaluation schemes}
\label{sec:eval}

Before solving a problem, one must define what a correct solution looks like. In machine learning, this is typically achieved by calculating some metrics, such as precision (PRC) and recall (RCL), with an \textit{evaluation scheme} or algorithm. These metrics attempt to reflect, in a somewhat subjective sense, the true model performance in a given task. In object detection, a traditional choice for an evaluation scheme is the scoring algorithm from PASCAL VOC2012 challenge \cite{Everingham2011}, which mainly uses the average precision (AP) metric. However, the performance metric calculation depends on first classifying all predictions under evaluation into either true positive (TP) or false positive (FP) category given the ground truth (GT) data labels.

In object detection, this classification is traditionally done by a \textit{ground truth matching algorithm}, such as \cref{algo:generic}. It is a key design choice in an object detection evaluation scheme implementation, as it decides what types of predictions are rewarded (marked as TPs) and what are penalized (marked as FPs). For example, duplicate detections are typically marked as FPs by most evaluation schemes, such as the VOC2012 algorithm \cite{Russakovsky2015}. What can be confusing is that most evaluation schemes report the same metrics (PRC, RCL, AP), albeit their way of arriving at those metrics is fundamentally different due to the choice of the ground truth matching algorithm. We next review a couple of existing evaluation schemes, discuss their issues in the SAP-APD context, and finally compare them to our proposed evaluation scheme.

The VOC2012 evaluation scheme is a widely used method for evaluating the performance of object detectors that output bounding box predictions and their associated confidence scores \cite{Everingham2010, Everingham2011, Lin2014, Russakovsky2015, Kuznetsova2020}. It is designed for relatively large objects and requires strict alignment of the detection and label box measured by the \textit{intersection over union} (IoU) metric given by \cref{eq:iou}. Therefore, VOC2012 evaluation uses the a typical TP criterion of over $0.5$ IoU between the two boxes \cite{Lin2014}. Moreover, each prediction must be matched with exactly one label (one-to-one mapping). In essence, our generalized \cref{algo:generic} yields GT matching for VOC2012 evaluation with the input parameters $\varepsilon=0.5$, $g_{\max}=1$ and $a_{\min}=0$. These strict requirements are the root cause of the dense group separation problem in HERIDAL, as they encourage prediction postprocessing methods that likely produce results similar to what is depicted in \cref{fig:nms}. On the other hand, results like in \cref{fig:mob} receive zero points (one FP and five FNs) under VOC2012 evaluation because of two reasons. Firstly, the IoU requirement is not satisfied with any of the labels, and secondly, no label grouping is allowed. Nonetheless, the MOB results in \cref{fig:mob} are arguably more informative and visually pleasing in the SAR visual search context.

The recent MS COCO evaluation scheme has largely replaced the aforementioned VOC2012 criteria due to the advent of the popular MS COCO object detection challenge \cite{Lin2014}. MS COCO evaluation is very similar to VOC2012, however, it computes some additional metrics on top of VOC2012 AP, such as iterating the evaluation algorithm over different IoU threshold $\varepsilon$ values from 0.5 to 0.95 and taking the mean AP as the end result \cite{Liu2020}. Thus, the MS COCO scheme is even stricter than VOC2012 in terms of object localization accuracy requirements, and is thereby far too strict for the SAR-APD problem evaluation. Consequently, we omit the detailed discussion of the MS COCO evaluation in this paper, and use the VOC2012 scheme as a baseline for comparison instead. 

\begin{figure}[t]
	\begin{center}
		\includegraphics[width=0.6\linewidth]{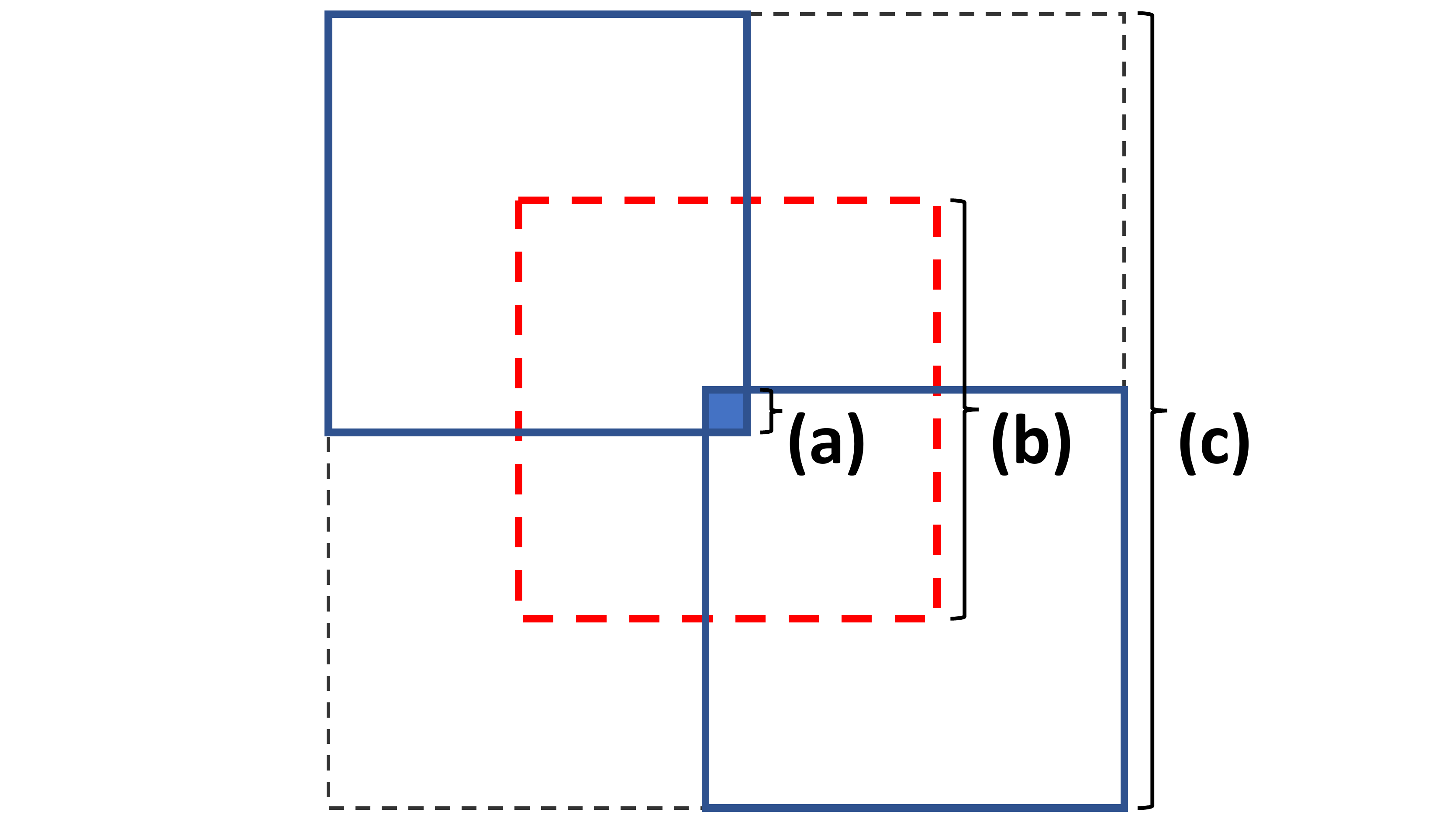}
		\caption{Illustration of localization area limits given an accuracy requirement defined by a fixed evaluation IoU threshold $\varepsilon$: (a) average human width $w_{\text{avg}}^h$ in aerial images, (b) maximum width for prediction bounding box $w_{\max}^p(\varepsilon)$, and (c) maximum acceptable localization area width $w_{\max}^{\text{TP}}(\varepsilon)$ for a maximum-sized prediction box. See equations \eqref{eq:iou-thres} and \eqref{eq:iou-thres-res} for calculating the values of $w_{\max}^p(\varepsilon)$ and $w_{\max}^{\text{TP}}(\varepsilon) $ respectively. Note that only prediction boxes with unity aspect ratio are considered here for simplicity.}
		\label{fig:loc-accuracy}
	\end{center}
\end{figure}

The proposed SAR-APD evaluation scheme matches ground truth boxes such that the real-world, loose localization requirement of aerial SAR is respected. Hence, we \textit{ignore} the dense group separation problem by allowing one-to-many mappings between a prediction and label boxes. Furthermore, we significantly relieve the evaluation IoU threshold $\varepsilon$ from 0.5 to 0.0025, which allows much larger prediction boxes to be considered as TPs (see \cref{fig:loc-accuracy}). As such, the red predicted box in \cref{fig:mob} would achieve five TPs under SAR-APD evaluation, as this single prediction captures all the blue label boxes. The ground truth matching method in our SAR-APD evaluation scheme corresponds to \cref{algo:generic} with the input parameters $\varepsilon=0.0025$, $g_{\max}=\infty$ and $a_{\min}=0.25$. Although, in our experiments we used $a_{\min}=0$. We argue this limit on box minimum size had negligible effect on our performance metrics due to the MOB algorithm tendency of overestimating the bounding box size, as seen in \cref{fig:mob}. Nonetheless, with the relaxed $\varepsilon=0.0025$ value, the predicted box minimum size becomes an important evaluation scheme design consideration. That is, barely visible, few pixels wide predictions should not be accepted as TPs. Moreover, all parameters of our generalized \cref{algo:generic} can be customized for different use cases other than SAR-APD, for example by limiting the maximum prediction group size by changing $g_{\max}$.

\setlength{\textfloatsep}{4ex}{
	\begin{algorithm}[t]
		\footnotesize
		\caption{Generalized ground truth matching method for typical object detector performance evaluation.}\label{algo:generic}
		\vspace{1.5ex}
		\hspace*{\algorithmicindent}\hspace{0.18em}\textbf{Input:}\hspace{1.15em}$\mathcal{B}^p = \{(b_i^p, s_i)\}_{i=1}^D$\Comment{$D$ bounding box predictions sorted} \\ 
		\hspace*{\algorithmicindent} \Comment{by decreasing confidence score $s_i$} \\ 
		\hspace*{\algorithmicindent} \Comment{for class $c$ from input image $\mathbf{I}$.} \\ 
		\hspace*{\algorithmicindent}\hspace{0.25em}\phantom{\textbf{Input:}}\hspace{1.15em}$\mathcal{B}^g = \{b_k^g\}_{k=1}^N$\Comment{$N$ ground truth bounding box labels} \\ 
		\hspace*{\algorithmicindent} \Comment{for class $c$ from input image $\mathbf{I}$.} \\ 
		\hspace*{\algorithmicindent}\hspace{0.25em}\phantom{\textbf{Input:}}\hspace{2em}$\varepsilon \in [0, 1] \subset \mathbb{R} $\Comment{Box IoU threshold for matching.}  \\ 
		\hspace*{\algorithmicindent}\hspace{0.25em}\phantom{\textbf{Input:}}\hspace{0.25em}$g_{\max} \in \mathbb{N} $\Comment{Maximum number of GT boxes $b_k^g$}  \\ 
		\hspace*{\algorithmicindent} \Comment{to match with a single prediction $b_i^p$.} \\ 
		\hspace*{\algorithmicindent}\hspace{0.25em}\phantom{\textbf{Input:}}\hspace{0.375em}$a_{\min} \in [0, 1] \subset \mathbb{R} $\Comment{Minimum value for  $A(b_{k}^p)/A(b_{k}^g)$,}  \\ 
		\hspace*{\algorithmicindent} \Comment{which limits TP prediction box size.} \\
		\hspace*{\algorithmicindent}\hspace{0.18em}\textbf{Output:} \hspace{0.8em}$\mathcal{Y} \in \{0, 1\}^{X}$ \Comment{A binary sequence of variable length} \\
		\hspace*{\algorithmicindent} \Comment{$X \in \mathbb{N}_0$ indicating true and false } \\
		\hspace*{\algorithmicindent} \Comment{positives, if $g_{\max}=1 \Rightarrow X=D$.} \\ \vspace{-1.5ex}
		\begin{algorithmic}[1]
			\Function{MatchBoxesGeneric}{$\mathcal{B}^p, \mathcal{B}^g, \varepsilon, g_{\max}, a_{\min} $}
			\State $\mathcal{Y} \gets \emptyset$ 
			\For{$i \gets 1, \hdots, D$} \Comment{Iterate over predictions.}
			\State $ \mathcal{M} \gets \emptyset$ \Comment{Initialize matched label set.}
			\State $ \mathbf{t} \gets \operatorname{IoU}(b_i^p, \mathcal{B}^g$)  \Comment{GT IoU value vector (see \cref{eq:iou}).}
			\State $ \operatorname{sort}(\mathcal{B}^g, \operatorname{argsort}(\mathbf{t}))$ \Comment{Sort $\mathcal{B}^g$ by descending IoU value.}
			\State $ \operatorname{sort}(\mathbf{t})$ \Comment{Sort $\mathbf{t}$ into descending order.}
			\For{$k \gets 1, \hdots, |{B}^g|$} 
			\State $S \gets A(b_{i}^p) > a_{\min} A(b_{k}^g)$
			\If{$t_{k} \geq \varepsilon$ and $|\mathcal{M}| < g_{\max}$ and $S$}
			\State $ \mathcal{M} \gets \mathcal{M} \cup \{b_{k}^g\} $ 
			\State $\mathcal{Y}  \gets  \mathcal{Y} \cup \{1\}$ \Comment{Attribute one TP to this prediction.}
			\EndIf
			\EndFor
			\If{$|\mathcal{M}| = 0$}
			\State $\mathcal{Y} \gets \mathcal{Y} \cup \{0\}$ \Comment{Nothing matched, add one FP.}
			\EndIf
			\State $ \mathcal{B}^g \gets \mathcal{B}^g - \mathcal{M} $ \Comment{Remove matched labels.}
			\EndFor
			\State \Return{$\mathcal{Y}$}
			\EndFunction
		\end{algorithmic}
	\end{algorithm}
}

The proposed SAR-APD evaluation scheme encourages results that effectively and reliably summarize the relevant, mission-critical information for SAR operations without much clutter. This is especially important if the SAR-APD algorithm co-operates with a human supervisor, which is the most likely scenario given the potential dire consequences of any mistakes. The reason is that any unnecessary distraction or irritation in a stressful SAR situation can undermine the actions and decisions made by the said supervisor. For example, too many false positives can cause the SAR-APD algorithm suggestions to be mistrusted and ignored.

Moreover, our evaluation scheme enables more research effort to be directed towards more meaningful, value-creating endeavors, such as ensuring high true positive rate for these approximate group labels or improving the detector processing speed, instead of spending excessive time tweaking threshold parameters to near perfectly delineate each tiny person. In addition, the proposed SAR-APD evaluation scheme works as a simple drop-in replacement for the VOC2012 criteria without the need to modify any data labels. In fact, the SAR-APD evaluation scheme can define implicit group labels with configurable size for typical object detection data sets.



The relationship between the evaluation IoU threshold $\varepsilon=0.0025$ and maximum acceptable pixel width for the prediction bounding box $w_{\max}^p(\varepsilon)$ can be formulated as

\begin{equation}
	\label{eq:iou-thres}
	w_{\max}^p(\varepsilon) =  \frac{ w_{\text{avg}}^h }{ \sqrt{\varepsilon} } = \frac{60}{\sqrt{0.0025}}= 1200, \\
\end{equation}

where $w_{\text{avg}}^h$ is the estimated average human pixel width in the HERIDAL aerial images. Note that this equation only applies when the ground truth bounding box is completely inside the predicted box, otherwise $w_{\max}^p(\varepsilon)$ will be lower. This scenario is depicted in \cref{fig:loc-accuracy}. Furthermore, \cref{fig:loc-accuracy} illustrates the maximum acceptable (unity aspect ratio) area width $w_{\max}^{\text{TP}}(\varepsilon) $ for prediction box of width $w_{\max}^p(\varepsilon)$ to exist in, that is given by

\eqvskip
\begin{equation}
	\label{eq:iou-thres-res}
	w_{\max}^{\text{TP}}(\varepsilon) = 2w_{\max}^p(\varepsilon) - w_{\text{avg}}^h = 2\cdot 1200-60 = 2340.
\end{equation}

Therefore, a square prediction box of width $w_{\max}^p(\varepsilon)$ is only considered a TP inside this square region of width $w_{\max}^{\text{TP}}(\varepsilon)$ centered at the GT label, as shown in \cref{fig:loc-accuracy}. In other words, $w_{\max}^{p}=1200$ pixels corresponds to roughly 20 meters on the ground in most HERIDAL images \cite{Bozic-Stulic2019}. This means the SAR-APD evaluation requires the person to be located within the maximum predicted 20 m $\times$ 20 m square area on the ground, which is still in alignment with the localization accuracy recommendation by \etalcite{Molina}{Molina2012}. Hence, the SAR-APD evaluation scheme scores object detection algorithms based on their ability to find approximate georeferences for the objects of interest.


\section{MOB postprocessing}
\label{sec:mob}

In this work, we also propose a novel postprocessing method for the SAR-APD problem, called the MOB algorithm. It is a probabilistic BBA method that trades localization accuracy for detection certainty. MOB first clusters prediction boxes based on an overlap distance metric given by \cref{eq:jaccard}. Subsequently, MOB merges all candidate boxes in each overlap cluster into a single enclosing box by using \cref{eq:enclosing-box}. This is in contrast to the \textit{de facto} NMS method that attempts to eliminate all sufficiently overlapping boxes but the max-scoring one in a greedy manner. It could be argued that the popularity of NMS is largely based on its remarkable synergy with the VOC2012 evaluation method that emphasizes accurate individual object delineation, which is not the goal of SAR-APD. Conversely, MOB (with its default merge strategy) is not compatible with the VOC2012 method at all, as explained in \cref{sec:eval}. Therefore, we use our proposed SAR-APD scheme for measuring its performance truthfully. However, this makes the fair comparison of MOB to other BBA methods difficult at the moment.


Essentially, MOB helps the SAR personnel to focus on the relevant clues in the image, if there are any. This is achieved by eliminating clutter from the detection results and highlighting broad regions that include all the relevant objects of interest, as done in \cref{fig:mob}.

MOB is based on the single linkage agglomerative clustering algorithm implemented in the \texttt{scikit-learn} machine learning library \cite{scikit-learn}, which uses the Jaccard distance metric $d_j \in [0,1] \subset \mathbb{R}$ \cite{jaccard}. This metric and the corresponding pair-wise distance matrix $\mathbf{D}^j$ are defined as

\eqvskip
\begin{align}
	\label{eq:jaccard}
	\mathbf{D}_{ik}^j = d_j(b_i, b_k) &= 1 - \text{IoU}(b_i, b_k), \\
	\label{eq:iou}
	\operatorname{IoU}(b_i, b_k) &= \frac{A(b_i \cap b_k)}{A(b_i \cup b_k)}, 
\end{align}

where $b_i$ and $b_k$ are two input bounding boxes, and $A(b)$ denotes the area of region $b \subset \mathbb{R}^2$. Intuitively, $ d_j(b_i, b_k) = 1$ when $b_i$ and $b_k$ have zero overlapping area, and $ d_j(b_i, b_k) = 0$ when they are the same box (100\% overlap), as is the case with the diagonal elements of $\mathbf{D}^j$.

The BBA IoU threshold $\omega$ controls whether two boxes are considered overlapping. Its default value is $\omega=0.5$ in NMS. For MOB, we set $\omega=0$, which is experimentally found to result in the merging of all overlapping bounding boxes. This clustering formulation is equivalent to finding all the connected components in a Jaccard distance graph defined by $\mathbf{D}_{ik}^j$, if $ d_j(b_i, b_k) = 1-\omega = 1$ is considered an infinite distance, and thus there is no link between the bounding boxes $b_i$ and $b_k$.

After forming the overlapping bounding box clusters, we calculate the enclosing box for each cluster according to the \textit{enclose merge strategy}, given by \cref{eq:enclosing-box}. This sequence of enclosing boxes is the output of one \textit{MOB iteration} (i.e., one pass of the MOB algorithm). Particularly, consider a $C$-sized sequence of axis-aligned bounding boxes $\mathcal{B} = \{(\alpha, \beta)_i\}_{i=1}^C$ in one overlap cluster, where each box $(\alpha, \beta)_i$ is defined by its minimum $\alpha \in \mathbb{N}_0^2$ and maximum coordinate $ \beta \in \mathbb{N}_0^2$. Consequently, the enclosing bounding box $b_{\text{enclose}}$ of the cluster can be computed as

\eqvskip
\begin{align}
	\label{eq:enclosing-box}
	b_{\text{enclose}} &= (\alpha_{\min},  \beta_{\max}), \\
	\label{eq:a-min}
	\alpha_{\min} &= (\min \{\alpha_x \mid \alpha \in \mathcal{B}_\alpha\}, \min \{\alpha_y \mid \alpha \in \mathcal{B}_\alpha\}),  \\
	\label{eq:b-min}
	\beta_{\max} &= (\max \{\beta_x \mid \beta \in \mathcal{B}_\beta\}, \max \{\beta_y \mid \beta \in \mathcal{B}_\beta\}),
\end{align}

where $\mathcal{B}_\alpha = \{\alpha_i\}_{i=1}^C $ and $\mathcal{B}_\beta = \{\beta_i\}_{i=1}^C $. Moreover, the new confidence score of the enclosed box is computed as the mean of predicted box scores in the overlap cluster. This can sometimes yield relatively low confidence scores for the resulting boxes, so MOB also offers a \textit{top-k} parameter that can be used to prune some of the low confidence outlier boxes in a overlap cluster. This pruning frequently reduces the output box size as well.

On the other hand, sometimes this enclose merge strategy, given by \cref{eq:enclosing-box}, creates new, larger overlapping boxes. Because of this it can be desirable to repeat the MOB iteration a few times to ensure everything gets merged together. The number of these MOB iterations can be specified with an algorithm input parameter $M_{\max}$. However, if there is only one merged bounding box left after any MOB iteration, the algorithm terminates early. In our experiments, we use $M_{\max}=3$, which is experimentally found to be sufficient for merging all overlapping bounding boxes in most cases.

However, sometimes the merged boxes can grow undesirably large, especially if the original overlapping boxes were organized in a diagonal formation. For this reason, MOB allows one to specify a \textit{maximum inflation factor} $I_{\max}$, which limits the size of merged bounding boxes. It sets an upper bound for the result box areas that is given by 

\eqvskip
\begin{equation}
	\label{eq:inflation-factor}
	A_{\max} = I_{\max} \max \{A(b) \mid b \in \mathcal{B} \},
\end{equation}

where $\mathcal{B}$ is a sequence of input bounding boxes to the MOB algorithm, an example of which is illustrated in \cref{fig:no-bbox-aggregation}. In our experiments we use $I_{\max} = 100$, which means no output box can be larger than 100 times the largest input box size. As such, if the area of any merged bounding box exceeds $A_{\max}$ during MOB clustering, its underlying box cluster is recursively subdivided into smaller box clusters until $A_{\max}$ is no longer exceeded by any merged box. The subdivision algorithm uses a simple heuristic rule that divides a box cluster into two subclusters along the maximum length axis of the box cluster, such that each subcluster has a roughly equal number of boxes in them. This functionality limits the size inflation of the merged boxes. It is useful if a certain localization accuracy needs to be met, such as the one given by $\varepsilon$ in \cref{algo:generic}. Using $I_{\max} < (w_{\max}^p(\varepsilon))^2 / (w_{\text{avg}}^h)^2  < 1200^2/ 60^2 < 400$ guarantees that the localization accuracy requirement of the proposed SAR-APD evaluation scheme is satisfied.

The most obvious improvement in MOB is the improved visual representation for the dense people group predictions, as shown in \cref{fig:mob}. Such groups are realistic, yet rare, occurrences in SAR \cite{Bozic-Stulic2019, Goodrich2008}. MOB also makes prediction more robust by simplifying the SAR-APD problem. That is, MOB is far less sensitive to the choice of various threshold parameters, such as $\omega$, in terms of precision (when using SAR-APD evaluation). This enables the use of considerably lower threshold values that can increase the recall metric, as depicted in \cref{fig:calibration-study}. Therefore, fewer search targets are likely to go unnoticed during aerial drone searches. Moreover, there is less need for threshold parameter calibration on new unseen data, which makes MOB a more general approach to SAR-APD than NMS.

Finally, it should be mentioned that the enclose method given by \cref{eq:enclosing-box} is only one potential merge strategy. MOB also supports other similar strategies, such as calculating the cluster average box, to further customize its behavior for different potential use cases.


\section{AIR detector}
\label{sec:air}

Next, we briefly describe our approach to this SAR-APD problem, namely the AIR detector \thesiscite. It is built on top of the famous one-stage RetinaNet detector by \etalcite{Lin}{Lin2017Retina}. The AIR implementation is based on the \texttt{keras-\allowbreak retinanet} Python framework by Gaiser \cite{keras-retinanet}. 


The first aspect in designing AIR is naturally to address the \textit{small object detection} problem, an instance of which SAR-APD essentially is. For this, the built-in feature pyramid network (FPN) \cite{Lin2017Feature} of RetinaNet is essential, since it enhances the semantic quality of high-resolution CNN feature maps that capture the small objects. Moreover, the novel \textit{focal loss} used in training the RetinaNet detector helps to mitigate the severe foreground-background class imbalance in the HERIDAL data \cite{Lin2017Retina}. According to \etalcite{Lin}{Lin2017Retina}, the focal loss is shown to solve this imbalance problem better than various sampling heuristics, such as OHEM \cite{Shrivastava2016}. Furthermore, we employ a variant of image tiling strategy \cite{Unel2019} to enable the full pixel information usage in 4K input images within typical GPU memory limits.

Secondly, one needs to solve the training data scarcity problem of HERIDAL. Therefore, we use CNN backbone weights pretrained on ImageNet1000 data \cite{Russakovsky2015} for all our AIR experiments. This is a common strategy to mitigate the lack of big data in learning low-level feature presentations necessary for most computer vision tasks \cite{Shorten2019, Sun2017, Zoph2019}. In addition, we adopt the online, random data augmentation scheme by \etalcite{Zoph}{Zoph2019} with slight modifications, such as adding white Gaussian noise to the images. The idea is to incorporate label invariance to small geometric perturbations and lighting changes into the training data \cite{Shorten2019} by augmenting it with various geometric transformations and color operations. Furthermore, random data augmentation can regularize model training as well \cite{Krizhevsky2009}.

Lastly, we naturally employ the novel MOB algorithm, discussed in \cref{sec:mob}, for robust AIR output postprocessing during test time. However, we also use conventional NMS postprocessing as a performance benchmark, and for comparing our results to other notable SAR-APD work in \cref{tab:comparison-sota}.


\section{Experiments and results}
\label{sec:experiments}

\begin{figure}
	\centering
	\captionsetup[subfigure]{labelfont=bf}
	\begin{subfigure}{\linewidth}
		\centering
		\includegraphics[width=\linewidth]{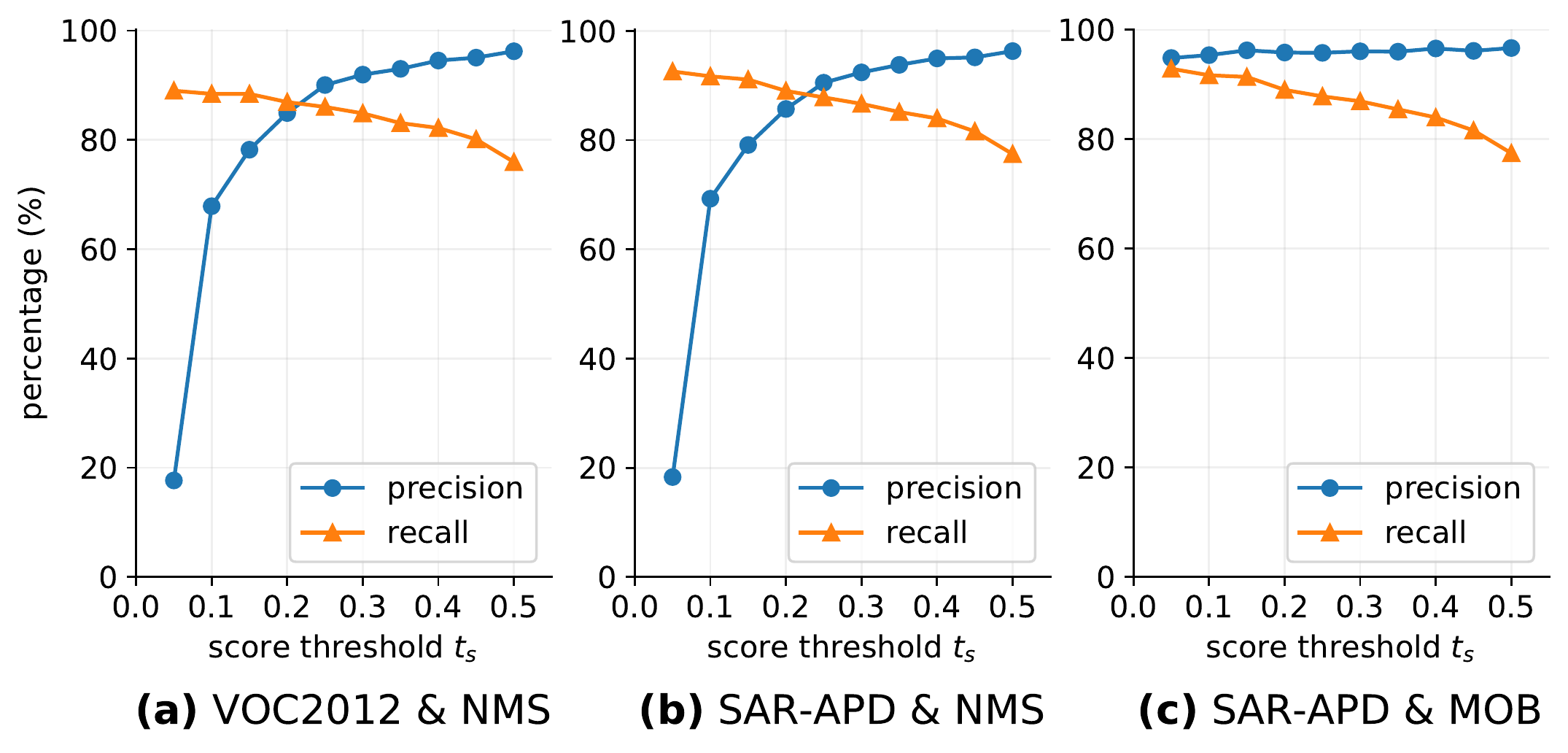}
		\refstepcounter{subfigure}
		\label{fig:calib-a}
	\end{subfigure}%
	\begin{subfigure}{0.0\linewidth}
		\centering
		\refstepcounter{subfigure}
		\label{fig:calib-b}
	\end{subfigure}%
	\begin{subfigure}{0.0\linewidth}
		\centering
		\refstepcounter{subfigure}
		\label{fig:calib-c}
	\end{subfigure}%
    \vspace{-4.0ex}
	\caption{AIR detector model calibration study results on the HERIDAL test set with different evaluation and bounding box aggregation methods. The choice of score threshold calibration parameter $t_s$ yields different trade-offs between precision and recall.}
	\label{fig:calibration-study}
\end{figure}

A total of six major experiments were conducted in order to create the AIR detector: a general hyperparameter search, backbone architecture selection (see \cref{tab:backbone-selection}), ablation study (see \cref{tab:ablation-study}), random trial experiment, model calibration study (see \cref{fig:calibration-study}), and a tiling performance study. For full details of these experiments, see the thesis work by \thesiscitelong. In this paper, we focus on the most relevant results that can help to improve SAR-APD performance in general.

\begin{table}
	\centering
	\tablestyle{4.0pt}{1.1}
	\begin{tabularx}{\linewidth}{l *{3}{Y} *{3}{Y}}
		\toprule
		\multicolumn{1}{c}{} & \multicolumn{3}{c}{Train set results (\%)} & \multicolumn{3}{c}{Test set results (\%)} \\ 
		\cmidrule(lr){2-4} \cmidrule(lr){5-7}
		Backbone
		& PRC  & RCL  & AP 
		& PRC  & RCL  & AP \\  
		\midrule
		\rowcolor{highlightrow}
		VGG16 \cite{Simonyan2015}
		& 74.4 & \textbf{98.9} & \textbf{89.1} & 79.8 & 79.8 & 76.5 \\
		ResNet50 \cite{He2016}
		& 74.6 & 91.8 & 77.8 & 83.5 & 72.1 & 69.6 \\
		\rowcolor{highlightrow}
		ResNet101 \cite{He2016}
		& 76.9 & 94.9 & 77.4 & 79.3 & 75.1 & 72.3 \\
		\textbf{ResNet152} \cite{He2016}
		& \textbf{81.8} & 95.8 & 81.7 & \textbf{85.0} & \textbf{80.7} & \textbf{78.1} \\
		\rowcolor{highlightrow}
		DenseNet201 \cite{Huang2017}
		& 77.7 & 42.6 & 35.4 & 73.9 & 28.5 & 26.3 \\
		SeResNeXt101$^{\dagger}$ \cite{Hu2018}
		&  -      & -      & -        & 77.6 & 69.7 & 65.8 \\
		\rowcolor{highlightrow}
		SeResNet152 \cite{Hu2018}
		& 78.8 & 84.3 & 70.8 & 76.9  & 66.2 & 63.1 \\
		EfficientNetB4 \cite{Tan2019}
		& 66.9 & 82.2 & 57.9 & 79.5 & 58.8 & 54.5 \\
		\bottomrule
	\end{tabularx}
	\vspace{-1ex}
	\caption{CNN backbone architecture selection experiment results on the HERIDAL data set. ResNet152 is the clear winner. $^{\dagger}$ Train set evaluation was not possible due to a corrupted model file.}
	\label{tab:backbone-selection}
\end{table}

One of the most important experiments is the CNN backbone selection for SAR-APD, since the choice of a feature extractor is a fundamental aspect of solving any object detection problem \cite{Jung2018, deeplearn, Liu2020}. Therefore, we test fine-tuning eight different ImageNet pretrained backbone architectures on HERIDAL, and the results are shown in \cref{tab:backbone-selection}. Here, we report the PRC, RCL and AP metrics computed using the VOC2012 evaluation scheme. As evident, ResNet152 outperforms all the other candidates on the test set, thus, it is selected as the CNN backbone for later experiments. Surprisingly, the simplest VGG16 network yields comparable results to ResNet152, while the last four most sophisticated models in \cref{tab:backbone-selection}, according to ImageNet classification score \cite{Liu2020}, perform relatively poor on HERIDAL.

\begin{table}[t]
	\centering
	\tablestyle{0.0pt}{1.1}
	\begin{tabularx}{\linewidth}{r *{12}{Y}}
		\toprule
		& \multicolumn{9}{c}{Ablations} & \multicolumn{3}{c}{Test results (\%)} \\ 
		\cmidrule(lr){2-10} \cmidrule(lr){11-13}
		& (a) & (b) & (c) & (d) & (e) & (f) & (g)  & (h)  & (i) & PRC & RCL & AP \\  
		\rowcolor{highlightrow}
		\midrule
		\rowlabel{1}{tab:ablation-study:row:empty}        
		& & & & & & & & & &                                                                                                                                      57.7 &    4.5 &  2.9 \\
		\rowlabel{2}{tab:ablation-study:row:a}
		& \faCheck & & & & & & & & &                                                                                                                       76.5 &  57.9 & 52.2 \\
		\rowcolor{highlightrow}
		\rowlabel{3}{tab:ablation-study:row:a+b}
		& \faCheck & \faCheck & & & & & & & &                                                                                                        87.8  & 72.4 & 69.3 \\
		\rowlabel{4}{tab:ablation-study:row:a+b+c}
		& \faCheck & \faCheck & \faCheck & & & & & & &                                                                                         85.0 & 80.7 & 78.1 \\
		\rowcolor{highlightrow}
		\rowlabel{5}{tab:ablation-study:row:a+b+c+d}
		& \faCheck & \faCheck & \faCheck & \faCheck & & & & & &                                                                          82.8  & 81.3 & 77.7 \\
		\rowlabel{6}{tab:ablation-study:row:a+b+c+e}
		& \faCheck & \faCheck & \faCheck &                 & \faCheck & & & & &                                                          57.7  & 84.3 & 81.7 \\
		\rowcolor{highlightrow}
		\rowlabel{7}{tab:ablation-study:row:a+b+c+e+f}
		& \faCheck & \faCheck & \faCheck &                 & \faCheck & \faCheck & & & &                                           17.2 & 81.9 & 79.2 \\
		\rowlabel{8}{tab:ablation-study:row:a+b+c+e+g}
		& \faCheck & \faCheck & \faCheck &                 & \faCheck &                 & \faCheck & & &                           49.5  & 79.5 & 77.3 \\
		\rowcolor{highlightrow}
		\rowlabel{9}{tab:ablation-study:row:a+b+c+e+h}
		& \faCheck & \faCheck & \faCheck &                 & \faCheck &                 &    & \faCheck & & 17.7 & \textbf{89.0} & \textbf{86.9} \\
		\rowlabel{10}{tab:ablation-study:row:a+b+c+e+h+i}
		& \faCheck & \faCheck & \faCheck &               & \faCheck &                 &    & \faCheck & \faCheck & \textbf{90.1} & 86.1 & 84.6 \\
		\bottomrule
	\end{tabularx}
	\vspace{-1ex}
	\caption{Ablation study results with the selected ResNet152 CNN backbone. The HERIDAL test set performance metrics are evaluated for each different type of added feature (ablation): (a) anchor boxes optimized for small objects, (b) improved hyperparameters, (c) image tiling, (d) pretraining on Stanford Drone data set \cite{Robicquet2016}, (e) online data augmentation pipeline, (f) decreased training example IoU threshold, (g) removing the two topmost layers from FPN, (h) best seed from random trial experiment and (i) calibrated score threshold parameter from model calibration study. The final model uses a combination of six ablations out of the nine tested.}
	\label{tab:ablation-study}
\end{table}

With the CNN backbone selected, we test several other tricks (or ablations) to improve AIR test performance, which are collected in \cref{tab:ablation-study}. We only keep the tested trick if it improves HERIDAL test AP metric, otherwise it is discarded from further experiments. The last calibration study ablation is an exception to this rule, as it primarily optimizes PRC instead.
\Cref{tab:ablation-study} includes our most important modifications for improving SAR-APD performance: (a) adjusting image region sampling grid, (b) tuning hyperparameters (e.g., learning rate), (c) using image tiling instead of resizing, (e) using data augmentation, and interestingly (h) finding appropriate random initialization for training.

Due to PRC sensitivity of NMS, it is beneficial to perform confidence score threshold parameter $t_s$ calibration as the last experiment (see \cref{tab:ablation-study}). To accomplish this, we test AIR with 10 different $t_s$ values from 0.05 to 0.5 on both VOC2012 and SAR-APD evaluation schemes. Moreover, we calibrate both the NMS and MOB postprocessing method, and the results are shown in \cref{fig:calibration-study}. For a balanced trade-off between PRC and RCL, we choose $t_s=0.25$ for NMS and $t_s=0.05$ for MOB.

Given the similarity of \cref{fig:calib-a} and \cref{fig:calib-b}, we can concur the choice of evaluation method affects the NMS-based detector relatively little in this experiment. On the other hand, by comparing  \cref{fig:calib-b} and \cref{fig:calib-c}, it is evident that MOB is far less sensitive to the choice of $t_s$ when compared to NMS. This is a compelling property of the MOB algorithm, which can in some cases eliminate the need for configuring $t_s$ parameter altogether. Furthermore, MOB enables setting $t_s$ much lower to increase recall without sacrificing precision significantly.

\begin{figure}
	\centering
	\captionsetup[subfigure]{labelfont=bf}
	\begin{subfigure}{\linewidth}
		\centering
		\includegraphics[width=\linewidth]{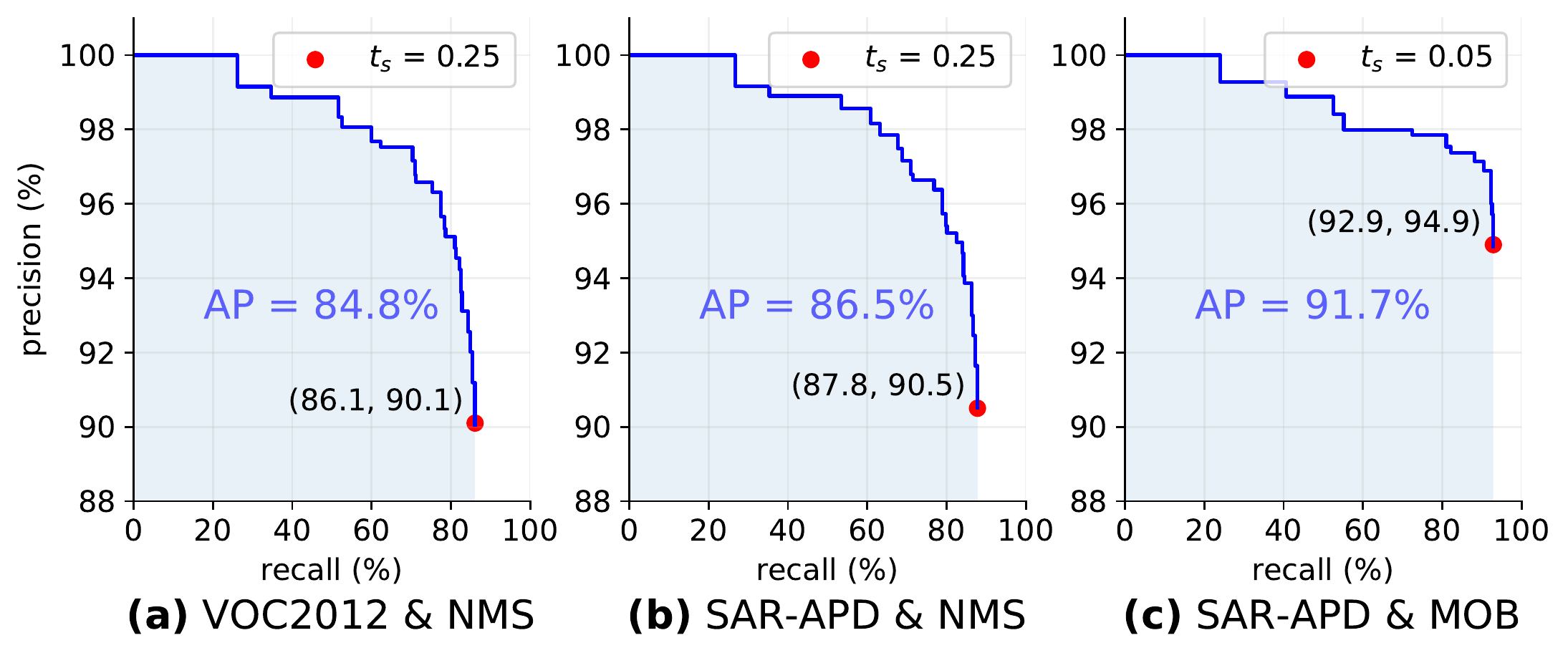}
		\refstepcounter{subfigure}
		\label{fig:pr-a}
	\end{subfigure}%
	\begin{subfigure}{0.0\linewidth}
		\centering
		\refstepcounter{subfigure}
		\label{fig:pr-b}
	\end{subfigure}%
	\begin{subfigure}{0.0\linewidth}
		\centering
		\refstepcounter{subfigure}
		\label{fig:pr-c}
	\end{subfigure}%
	\vspace{-4.0ex}
	\caption{Precision-recall curves for different evaluation experiments with the AIR detector on the HERIDAL test set. The red dot corresponds to a precision-recall pair at a certain fixed confidence score threshold value $t_s$. As such, by moving the dot along the curve, different trade-offs between precision and recall can be made. Notice that MOB enables moving the dot much further to the right without the precision metric plummeting. This encourages placing more emphasis on recall, which is critical for the SAR mission success rate.}
	\label{fig:pr-curves}
\end{figure}

With the score thresholds fixed, we analyze NMS and MOB performance further under both evaluation methods. For this, we plot precision-recall curves shown in \cref{fig:pr-curves}. Similar phenomenon to \cref{fig:calibration-study} can be seen here as well: SAR-APD evaluation affects NMS performance very little (see \cref{fig:pr-b}), likely mostly due to the relaxed $\varepsilon$ criterion. However, adding MOB to the equation significantly increases all measured performance metrics in \cref{fig:pr-c}. This can likely be attributed to the relaxation of the dense group separation problem, and thus to the lessened need for eliminating boxes with score thresholding to keep PRC satisfactory, which in turn is reflected on the increased RCL metric. This shows the importance of using accurate metrics.

\begin{table}
	\centering
	\tablestyle{1.0pt}{1.2}
	\begin{tabularx}{\linewidth}{l *{4}{Y}}
		\toprule
		Model
		&PRC (\%)&RCL (\%)&AP (\%)&ATI (s) \\  
		\midrule
		\rowcolor{highlightrow}
		\textit{VOC2012 evaluation} & & & & \\
		~~Mean shift clustering \cite{Turic2010}
		& 18.7 & 74.7 & -  & -\\
		\rowcolor{highlightrow}
		~~Saliency guided VGG16 \cite{Bozic-Stulic2019}
		& 34.8 & 88.9 & - & - \\
		~~Faster R-CNN (2019) \cite{Bozic-Stulic2019}
		& 58.1 & 85.0 & - & - \\
		\rowcolor{highlightrow}
		~~Faster R-CNN (2018) \cite{Marusic2018} 
		& 67.3 & 88.3 & \textbf{86.1} & \textbf{1} \\ 
		~~Multimodel CNN  \cite{Vasic2020} 
		& 68.9 & \textbf{94.7} & - & 10 \\ 
		\rowcolor{highlightrow}
		~~SSD  \cite{Vasic2020} 
		& 4.3 & 94.4 & - & - \\
		~~AIR with NMS (ours)
		& \textbf{90.1} & 86.1 & 84.6 & \textbf{1} \\ 
		
		\midrule
		
		\rowcolor{highlightrow}
		\textit{SAR-APD evaluation (ours)} & & & & \\
		~~AIR with NMS (ours)
		&  90.5 & 87.8 & 86.5 & \textbf{1} \\ 
		\rowcolor{highlightrow}
		~~AIR with MOB (ours)
		& \textbf{94.9} & \textbf{92.9} & \textbf{91.7} & \textbf{1} \\ 
		\bottomrule
	\end{tabularx}
	\vspace{-1ex}
	\caption{Comparison with the state-of-the-art in SAR-APD on the HERIDAL test set using two different evaluation methods. The abbreviation ATI refers to \textit{average time per image}, which is a very crude estimate due to software and hardware differences in the reported test environments \thesiscite. Under the traditional VOC2012 evaluation AIR achieves state-of-the-art results in both PRC and ATI. MOB boosts all accuracy metrics under SAR-APD evaluation.}
	\label{tab:comparison-sota}
\end{table}

The comparison of the AIR detector to the state-of-the-art SAR-APD methods is shown in \cref{tab:comparison-sota}. Both evaluation methods are included, however, the SAR-APD scheme results cannot be directly compared to other work. Our strongest competitor is the Multimodel CNN by Vasi\'{c} and Papi\'{c} \cite{Vasic2020}, which achieves a very high RCL. Nevertheless, this model is an order of magnitude slower due to its very high complexity (an ensemble model of four deep CNNs), as shown by its average time per image (ATI) metric in \cref{tab:comparison-sota}. This can hinder its practical adoption to SAR missions. Furthermore, it lacks end-to-end training capability. Some of the RCL gap can likely be attributed to better solving of the less important dense person group problem, as indicated by our MOB results in \cref{tab:comparison-sota}. As for the other strong competitor Faster R-CNN \cite{Marusic2018}, we argue AIR can achieve similar recall and AP while still being ${\sim}$11 points more precise by choosing $t_s=0.15$, as shown in \cref{fig:calib-a}.

Lastly, we highlight that our final MOB results in \cref{tab:comparison-sota} far exceed the human average performance in the same task. According to \thesiscitelong, the average human achieves roughly the following metrics in SAR-APD: 59\% PRC, 68\% RCL and 33 s ATI, with less than around 200 images to inspect. This is a fraction of the average, real-world number \cite{Gotovac2020}. Moreover, the human performance is demonstrated to drop with the number of images inspected due to accumulated fatigue \thesiscite. Therefore, the tireless AIR detector and its innovations can significantly improve drone-based SAR efficiency in the future.

\section{Conclusions}
\label{sec:conclusions}

The current aerial drone searches in SAR lack efficient means for visual footage inspection. Our deep learning solution to this problem showcased state-of-the-art performance (${\sim}$21 point increase in PRC) on the difficult HERIDAL benchmark. Moreover, we redefined the related SAR-APD problem by presenting a novel evaluation and BBA method for it. This is to better capture the real-world requirements of aerial SAR.
We also added extensive SAR-APD experimentation results to the scientific literature. Finally, we identified some important directions for future work as well. These include the use of MOB and SAR-APD evaluation for model validation during training, an extensive study of different MOB merge strategies, and the pursuit of achieving AIR real-time inference performance.


{\small
\bibliographystyle{ieee_fullname}
\bibliography{sources,paper-sources}
}

\end{document}